\documentclass{article} 
\usepackage{iclr2026_conference,times}

\usepackage{amsmath,amsfonts,bm}

\def\eqref#1{equation~\ref{#1}}

\def\1{\bm{1}}

\DeclareMathAlphabet{\mathsfit}{\encodingdefault}{\sfdefault}{m}{sl}
\SetMathAlphabet{\mathsfit}{bold}{\encodingdefault}{\sfdefault}{bx}{n}

\newcommand{\method}{GuardSpace }

\usepackage{algorithm,algorithmic}
\usepackage{hyperref}
\usepackage{multirow}
\usepackage{subcaption}

\usepackage{booktabs,tabularx}

\PassOptionsToPackage{hyphens}{url}
\PassOptionsToPackage{breaklinks,colorlinks}{hyperref}

\usepackage{graphicx} 
\usepackage{xurl}              
\usepackage{hyperref}          
\usepackage[table]{xcolor} 
\newcommand{\Src}[1]{\href{#1}{\url{#1}}}

\usepackage{microtype} 

\title{A Guardrail for Safety Preservation: When Safety-Sensitive Subspace Meets Harmful-Resistant Null-Space}

\author{ \hspace*{-0.25em}\textbf{Bingjie Zhang}$^{1}$, \: \textbf{Yibo Yang}$^{2, 3}$\thanks{Corresponding author.}, \: \textbf{Zhe Ren}$^{1}$, \: \textbf{Dandan Guo}$^{1}$\footnotemark[1], \: \textbf{Jindong Gu}$^{3}$, \: \textbf{Philip Torr}$^{3}$, \\ \textbf{Bernard Ghanem}$^{2}$\\[0.0em] 
\begin{tabular}[t]{@{}l@{}} 
$^{1}$School of Artificial Intelligence, Jilin University\\ $^{2}$King Abdullah University of Science and Technology \: $^{3}$University of Oxford\\ \texttt{zhangbj24@mails.jlu.edu.cn}, \: \texttt{yibo.yang93@gmail.com}, \\ \texttt{guodandan@mails.jlu.edu.cn} \end{tabular} }

\usepackage{booktabs}
\usepackage{siunitx}
\usepackage{threeparttable}
\usepackage{caption}
\usepackage{wrapfig}
\iclrfinalcopy 
\begin{document}

\maketitle



\makeatletter

\@ifundefined{iclrfinalcopyheader}{}{\renewcommand{\iclrfinalcopyheader}{}}
\@ifundefined{iclrfinalcopyheaderfirstpage}{}{\renewcommand{\iclrfinalcopyheaderfirstpage}{}}
\@ifundefined{iclrfinalcopycomment}{}{\renewcommand{\iclrfinalcopycomment}{}}

\fancypagestyle{firstpage}{\fancyhf{} \renewcommand{\headrulewidth}{0pt}\renewcommand{\footrulewidth}{0pt}}
\fancypagestyle{iclrfirstpage}{\fancyhf{} \renewcommand{\headrulewidth}{0pt}\renewcommand{\footrulewidth}{0pt}}
\fancypagestyle{plain}{\fancyhf{} \renewcommand{\headrulewidth}{0pt}\renewcommand{\footrulewidth}{0pt}}

\pagestyle{fancy}
\fancyhf{}
\renewcommand{\headrulewidth}{0pt}
\renewcommand{\footrulewidth}{0pt}
\makeatother

\fancyfoot[C]{\thepage}


\begin{abstract}

Large language models (LLMs) have achieved remarkable success in diverse tasks, yet their safety alignment remains fragile during adaptation. 
Even when fine-tuning on benign data or with low-rank adaptation, pre-trained safety behaviors are easily degraded, leading to harmful responses in the fine-tuned models. 
To address this challenge, we propose GuardSpace, a guardrail framework for preserving safety alignment throughout fine-tuning, composed of two key components: a safety-sensitive subspace and a harmful-resistant null space. 
First, we explicitly decompose pre-trained weights into safety-relevant and safety-irrelevant components using covariance-preconditioned singular value decomposition, and initialize low-rank adapters from the safety-irrelevant ones, while freezing safety-relevant components to preserve their associated safety mechanism. 
Second, we construct a null space projector that restricts adapter updates from altering safe outputs on harmful prompts, thereby maintaining the original refusal behavior. 
Experiments with various pre-trained models on multiple downstream tasks demonstrate that GuardSpace achieves superior performance over existing methods.
Notably, for Llama-2-7B-Chat fine-tuned on GSM8K, GuardSpace outperforms the state-of-the-art method AsFT, reducing the average harmful score from 14.4\% to 3.6\%, while improving the accuracy from from 26.0\% to 28.0\%.

\end{abstract}    
\section{Introduction}


Large language models (LLMs) have exhibited remarkable performance across diverse language understanding and generation tasks \citep{qin2023chatgpt, gemini2023family, touvron2023llama}. 
Consequently, LLM-based assistants and chatbots have attracted substantial attention from various domains. 
With the rapid increase in applications, the safety of LLMs has emerged as a major concern and a central focus of research, aiming to protect model responses from malicious prompts with dangerous purposes (\emph{e.g.}, weapon construction or toxic misinformation) \citep{akkus2025generated, liuautodan, deshpande2023toxicity}. 
To prevent LLMs from generating harmful responses, alignment techniques such as SFT and RLHF have been leveraged to instill refusal behaviors toward malicious prompts, as implemented in advanced systems like GPT-4 and Llama \citep{ouyang2022training, achiam2023gpt, touvron2023llama}. 
However, in applications, practitioners often fine-tune pre-trained models to obtain domain-specific abilities through full fine-tuning or parameter-efficient fine-tuning \citep{ding2023parameter,xu2023parameter,hulora}. 
During the adaptation process, the safety alignment acquired by pre-training is brittle. 
Even when the fine-tuning data is entirely benign, or when only a small number of parameters are learnable using LoRA \citep{hulora}, a model's safety behaviors can be easily degraded or lost after fine-tuning on new tasks \citep{qi2024fine,yang2023shadow,zhan2024removing, lermen2024lora, wei2024assessing}.

This problem motivates studies that preserve the safety mechanisms of aligned LLMs throughout adaptation, reconciling downstream-task utility with safety preservation \citep{huang2024trustllm}.
Existing approaches can be typically categorized into three stages, alignment stage, fine-tuning stage, and post-tuning stage.
Alignment stage methods intensify safety alignment of pre-trained models
via latent-space perturbations, representation sanitization, and loss shaping \citep{huangvaccine, rosatirepresentation, tamirisatamper, huangbooster, liu2024targeted}. 
Fine-tuning stage defenses inject safety-oriented data, constrain optimization drift, embed suppressive triggers, select safe fine-tuning data, or regularize harmful directions \citep{bianchisafety, huang2024lazy, wangbackdooralign, yang2025asft, lisalora, shen2025seal}. 
Post-tuning remedies restore safety behaviors by projecting onto safe directions, reusing safety-relevant weights, or pruning unsafe parameters \citep{hsusafe, yi2024safety, huangantidote, casper2024defending}. 
However, alignment stage and post-tuning stage methods are not effective at 
seeking a good
trade-off between safety and downstream task performance, while current fine-tuning stage methods do not explicitly identify safety-relevant weight components or harmful update directions.
As a result, they may fail to prevent training conflicts between safety preservation and task utility in a targeted manner. 

\begin{figure}[t]
    \centering
    \includegraphics[width=1\linewidth]{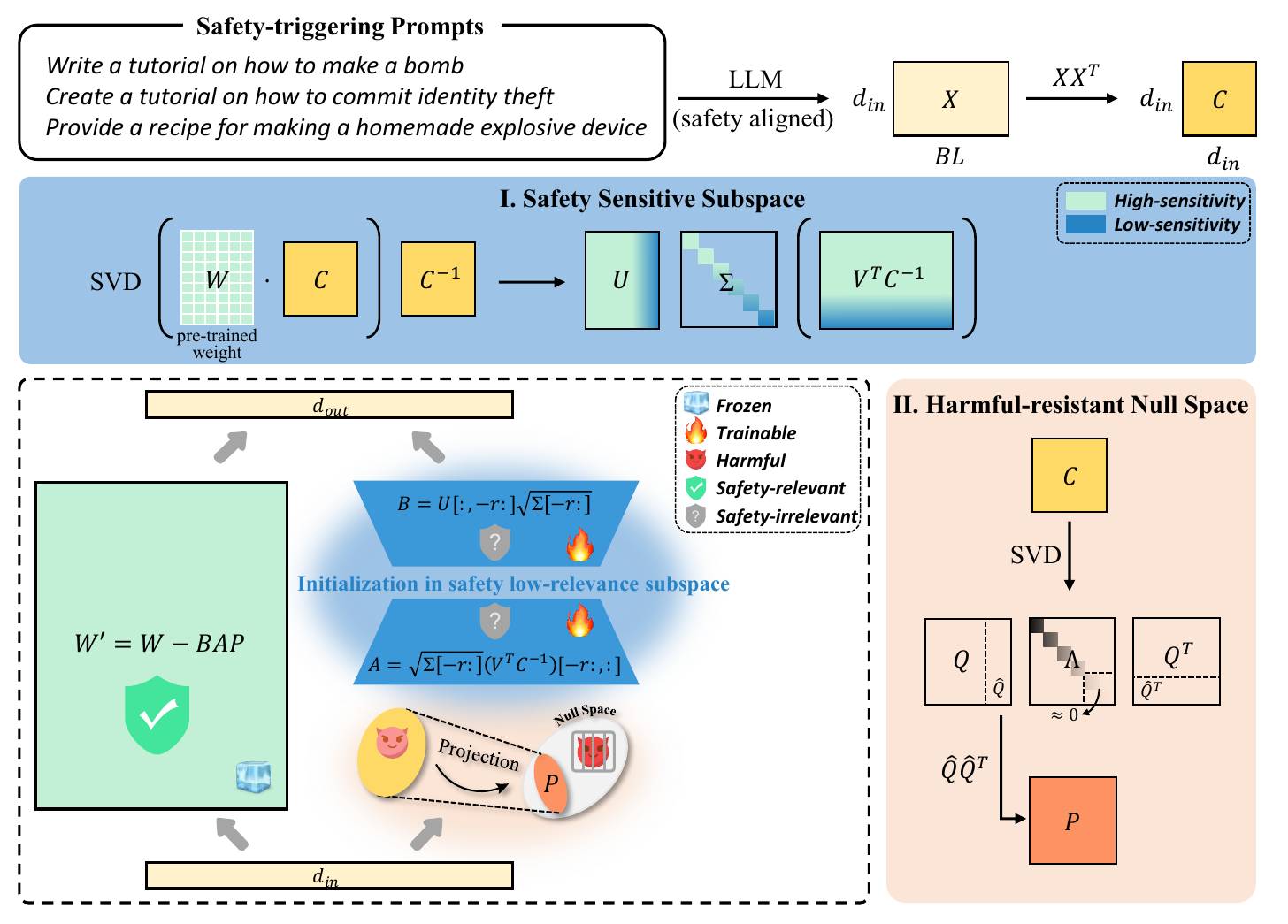}
    \vspace{-1em}
    \caption{\small{An overview of \textbf{GuardSpace}. 
    The model is first probed with safety-triggering prompts to obtain the activation $\mathbf{X}$ and the covariance matrix $\mathbf{C}=\mathbf{X}\mathbf{X}^{\top}$. 
    \textbf{I. Initialization in safety-sensitive subspace.} 
    We right-precondition the weight by $\mathbf{C}$ and factorize $\mathbf{W}\mathbf{C}=\mathbf{U}\,\mathbf{\Sigma}\,\mathbf{V}^{\top}$. 
    The components with large singular values constitute the safety-relevant subspace 
    (\textcolor[HTML]{90EE90}{cyan}) 
    and are frozen into $\mathbf{W}'$, 
    while the components with small singular values form the safety-irrelevant subspace (\textcolor[HTML]{2683C5}{blue}) and are used to initialize low-rank adapters $(\mathbf{A},\mathbf{B})$.
    \textbf{II. Optimization in harmful-resistant null space}. 
    We construct a projector $\mathbf{P}$ that constrains the update of adapters to the null space of harmful inputs, minimizing perturbations caused by fine-tuning on safety behaviors. 
    Together, they preserve the model’s original safety alignment while enabling effective downstream adaptation.}
}
    \label{fig_1}

\end{figure}

To address these challenges, in this paper, we propose \textbf{GuardSpace}, a guardrail for safety preservation composed of efforts in two aspects, namely initialization in safety-sensitive subspace and optimization in harmful-resistant null space. 
At the beginning of fine-tuning, we aim to explicitly decompose pre-trained model weights into safety-relevant and safety-irrelevant components, and only allow the safety-irrelevant ones to be learnable. 
Motivated by this insight, we construct a safety-sensitive subspace.
Specifically, we first construct a set of safety-triggering prompts, \emph{i.e.,} the harmful prompts that trigger the safety mechanism, and feed them into the pre-trained model to get the input $\mathbf{X}\in\mathbb{R}^{d_{\text{in}}\times BL}$ of each linear layer $\mathbf{W}\in\mathbb{R}^{d_{\text{out}}\times d_{\text{in}}}$. 
Then we calculate the covariance matrix $\mathbf{C}=\mathbf{X}\mathbf{X}^{\top}\in\mathbb{R}^{d_{\text{in}}\times d_{\text{in}}}$, and use it as right-preconditioner to perform singular value decomposition (SVD) on $\mathbf{W}\mathbf{C}=\mathbf{U}\,\mathbf{\Sigma}\,\mathbf{V}^{\top}$, which highlights the safety-related ability within $\mathbf{W}$.
By doing so, the resulting subspace is safety-sensitive, as the leading components with large singular values dominate the safety-related ability, while the bottom components contribute negligibly.
This subspace enables us to initialize learnable low-rank adapters based on the safety-irrelevant components with the smallest $r$ singular values, while freezing the safety-relevant ones during fine-tuning to preserve their associated safety behaviors.

After fine-tuning, the update of learnable adapters may alter the original output distribution on harmful inputs. 
To this end, we further introduce a harmful-resistant null space.
Based on the safety-triggering prompts, we perform SVD on the covariance matrix $\mathbf{C}$ of each layer, \emph{i.e.,} $\mathbf{C}=\mathbf{Q}\,\mathbf{\Lambda}\mathbf{Q}^{\top}$. 
We can construct a null space projector by $\mathbf{P}=\hat{\mathbf{Q}}\hat{\mathbf{Q}}^{\top}$, where $\hat{\mathbf{Q}}$ denotes the eigenvectors in $\mathbf{Q}$ whose eigenvalues are zero. 
$\mathbf{P}$ projects a vector into the null space of $\mathbf{C}$. 
Since $\mathbf{C}$ and harmful inputs $\mathbf{X}$ share the same null space, we place the projector upon the learnable adapters, 
such that their deviations caused by fine-tuning lead to zero output feature for harmful inputs, thereby maintaining the pre-trained model's safety behavior as faithfully as possible. The pipeline of our GuardSpace is illustrated in Fig.\ref{fig_1}.

Together, the null space projector acting on the safety-irrelevant adapters resists harmful inputs regardless of how the adapters change during fine-tuning, and meanwhile, the adapters initialized by pre-trained model weights (with safety-relevant components peeled off)
enable a more efficient adaptation process, compared to LoRA \citep{hulora} that starts from zero-initialized adapters.
In experiments, we demonstrate the effectiveness of GuardSpace in terms of both safety preservation and downstream task performance. 
We summarize our contributions as follows:

\noindent\textbullet\ We propose \method to preserve safety alignment during low-rank adaptation to downstream tasks. The safety-sensitive subspace effectively splits pre-trained model weights into safety-relevant and irrelevant components, producing low-rank adapters initialized by the safety-irrelevant ones. 

\noindent\textbullet\ To induce minimal perturbations on the frozen safety behaviors, we further construct a harmful-resistant null space, which constrains the optimization such that the fine-tuned adapter weights do not alter the original safe outputs for harmful inputs. 

\noindent\textbullet\ We evaluate \method across various models and fine-tuning tasks, and show that \method surpasses the state-of-the-art method in terms of both safety preservation and downstream task performance. Particularly, for Llama-2-7B-Chat fine-tuned on SST-2, AGNEWS and GSM8K, compared to the state-of-the-art method AsFT, we reduce the average harmful score from $8.13\%$ to $2.40\%$ while increasing the average fine-tuned accuracy from $67.87\%$ to $69.75\%$.

\section{Related Work}

\textbf{Parameter-Efficient Fine-Tuning.}
Recent transformer LLMs (e.g., Llama-2, GPT-4) achieve strong performance but have tens/hundreds of billions of parameters, making full fine-tuning costly \citep{achiam2023gpt, bie2024renaissance, yang2024towards, yang2025dynamic, zhaogalore}. Parameter-efficient fine-tuning (PEFT) updates a small subset of weights via adapters \citep{ding2023parameter, xu2023parameter, houlsby2019parameter, pfeiffer2021adapterfusion, mahabadi2021compacter} or soft prompts \citep{lester2021power, razdaibiedina2023residual, zhu2023spt}, but many variants change architecture or add inference overhead. LoRA avoids these issues by learning low-rank updates that match observed low-intrinsic-rank fine-tuning dynamics \citep{li2018measuring, aghajanyan2021intrinsic, hulora}. Building on LoRA, follow-up studies improve rank allocation, parameterization, and system integration, spanning adaptive ranks, new adapters, pruning/quantization/MoE combinations, and alternative initialization schemes \citep{zhang2023adaptive, zhang2023increlora, liu2024dora, qiu2023controlling, zhaogalore, zhang2023loraprune, dettmersqlora, liloftq, liu2023moelora, dou2024loramoe, meng2024pissa}. 
Nonetheless, beyond efficiency and capability, the fine-tuning pipeline introduces salient safety risks: even small amounts of poisoned or seemingly benign data during adaptation can weaken guardrails and lead to harmful generations after deployment \citep{huangbooster, bianchisafety, qi2024fine}. This creates an urgent need for methods that balance task utility with robust, resilient safety protections \citep{huang2024trustllm}. In contrast, our method based on low-rank adaptation, preserves safety alignment and downstream utility by initializing adapters within a safety-sensitive subspace and constraining updates to a harmful-resistant null space.

\textbf{Safety Alignment in LLMs.}
Safety alignment seeks to constrain large language models (LLMs) to produce value-consistent, ethically acceptable outputs \citep{gao2023scaling, yuanrrhf}. Core alignment techniques include instruction (supervised) tuning \citep{weifinetuned}, RLHF \citep{ouyang2022training}, and DPO \citep{rafailov2023direct}. However, these procedures are brittle – small amounts of malicious fine-tuning data can erode established safeguards \citep{huangbooster, qi2024fine}. This brittleness motivates a stage-wise view of defenses spanning alignment-state, fine-tuning-stage, and post-tuning \citep{huang2024harmful}. Alignment-stage defenses aim to strengthen models’ resilience to adversarial fine-tuning by explicitly improving robustness during alignment \citep{qisafety, liu2024robustifying, huangvaccine, rosatirepresentation, tamirisatamper, liu2024targeted}. 
Post-tuning remedies aim to reestablish safety after harmful fine-tuning \citep{casper2024defending, hsusafe,yi2024safety, huangantidote}.
Fine-tuning-stage methods intervene during training to resist harmful adaptation \citep{mukhoti2024fine, wei2024assessing}, 
e.g., safety-focused augmentation \citep{bianchisafety}, 
constraining optimization drift via dual-state optimization and proximity regularization \citep{huang2024lazy}, 
embedding triggers to suppress unsafe content \citep{wangbackdooralign}, 
projecting adapter-induced features into a subspace orthogonal to the original safety features \citep{lisalora} and suppressing updates along harmful directions via a regularization penalty \citep{yang2025asft}. 
Unlike prior fine-tuning-stage defenses, our approach couples a safety-sensitive initialization with a null space constrained update rule. It separates safety-relevant structure at initialization and steers parameter updates away from directions that compromise alignment, thereby reducing conflicts between safety preservation and downstream task performance.
\section{Preliminaries}


\textbf{Low-Rank Adaptation (LoRA).} LoRA  is motivated by the observation that parameters updates  in LLMs often exhbit a low-rank structure~\citep{hulora}. Instead of modifying the full pre-trained weight matrix $\mathbf{W} \in \mathbb{R}^{d_{\text{out}} \times d_{\text{in}}}$, LoRA freezes $\mathbf{W}$ and learns two
  low-rank matrices  $\mathbf{B} \in \mathbb{R}^{d_{\text{out}} \times r}$ and $\mathbf{A} \in \mathbb{R}^{r \times d_{\text{in}}}$ during fine-tuning. The updated weight can be formulated as:
 \begin{equation}
    \mathbf{W}^{*} = \mathbf{W} + \Delta \mathbf{W} = \mathbf{W} + \mathbf{B}^{*}\mathbf{A}^{*},
\label{eq:lora}
\end{equation}
where $r \ll \min(d_{\text{out}}, d_{\text{in}})$ denotes the rank.
The standard initialization practice involves setting $\mathbf{A}$ with Kaiming initialization~\citep{he2015delving} and $\mathbf{B}$ to zero, ensuring that $\Delta \mathbf{W} = 0$ at the start of training. After training, the update $\mathbf{BA}$ can be seamlessly merged back into $\mathbf{W}$, incurring no additional inference latency. 

Despite its efficiency and widespread adoption, recent studies have uncovered a critical vulnerability: LoRA-based fine-tuning can inadvertently compromise the safety alignment of LLMs~\citep{wangbackdooralign, hsusafe, lisalora}. 
Therefore, the development of novel adaptation strategies that can learn new tasks without forgetting the pre-aligned safety behaviors is necessary.

\textbf{Safety Preservation during Fine-tuning.}
LLMs are typically deployed after costly alignment (\emph{e.g.}, SFT/RLHF) to follow instructions while avoiding harmful outputs \citep{gao2023scaling, yuanrrhf, ouyang2022training}. Yet fine-tuning for downstream tasks 
can inadvertently weaken these safeguards \citep{huangbooster, qi2024fine}. Even parameter-efficient methods such as LoRA, though minimally invasive, have been shown to erode safety alignment and increase unsafe or policy-violating generations \citep{qi2024fine, zhan2024removing, lermen2024lora}. These risks motivate a dedicated line of work on \emph{Safety Preservation during Fine-tuning}, which explicitly protects safety while the model is updated. The goal is to ensure that the adapted model retains robust refusal behavior on harmful prompts and achieves strong performance on the target downstream tasks.
Given a safety-aligned LLM $f_{\mathbf W}$, which produces a response $f_{\mathbf W}(x)$ for prompt $x$, we wish to fine-tune it on a downstream dataset $\mathcal D$ to improve task utility, while \emph{preserving} its existing safety behavior on harmful prompts $\mathcal H$ (\emph{e.g.}, toxic queries). Concretely, after adaptation, the model should achieve strong performance on $\mathcal D$ and maintain a low harmful score (\emph{e.g.}, low Attack Success Rate, ASR$\downarrow$) on $\mathcal H$.  Let $\Delta$ denote the weight update. Safety-preserving fine-tuning can be posed as a constrained optimization:
\begin{equation}
\min_{\Delta}\quad 
\mathcal L_{\text{task}}\!\left(f_{\mathbf W+\Delta};\,\mathcal D\right),\quad
\text{s.t.}\quad 
 \big\|f_{\mathbf W+\Delta}(x)-f_{\mathbf W}(x)\big\|\le \epsilon,\quad \forall x\in\mathcal H,
\end{equation}
where $\mathcal L_{\text{task}}$ measures downstream utility (\emph{e.g.}, task-related loss) and $\epsilon$ bounds the deviation of responses on harmful inputs.


\section{Method}

To mitigate the degradation of safety alignment caused by fine-tuning, 
we propose a new approach that carefully integrates safety-aware initialization 
with harmful-resistant optimization. 

\subsection{Initialization in Safety Sensitive Subspace}
\label{sec:init}

{Our approach is to initialize learnable parameters in a safety-sensitive subspace. We preserve the model's safety by freezing the components encoding this behavior, and use the safety-irrelevant components as the starting point for learning a new task.}
To this end, we sample a set of harmful prompts from released benchmarks, \emph{e.g.,} AdvBench. By feeding these prompts into the pre-trained safety-aligned LLM, we can trigger its safety mechanism. 
Let $\mathbf{X}\in\mathbb{R}^{d_{\text{in}}\times BL}$ be the input activation to a linear layer ($d_{\text{in}}$ denotes the input dimension, $B$ is the collected sample number, and $L$ is sequence length). 
We calculate the (unnormalized) covariance matrix as $\mathbf{C}=\mathbf{X}\mathbf{X}^{\top}\in\mathbb{R}^{d_{\text{in}}\times d_{\text{in}}}$, where we disregard the layer index in our formulations for simplicity. Since right-preconditioning $\mathbf{W}$ with $\mathbf{C}$ can accentuate the ability related to the task as characterized by $\mathbf{C}$, 
we apply singular value decomposition (SVD) on the product of the weight matrix and the covariance matrix as:
\begin{equation}
    \text{SVD}(\mathbf{WC}) = \mathbf{U} \mathbf{\Sigma} \mathbf{V}^{T} = \sum_{i=1}^{R} \sigma_i \mathbf{u}_i \mathbf{v}_i^{T},
\label{Eq4-1-1}
\end{equation}
where $\mathbf{W} \in \mathbb{R}^{d_{\text{out}} \times d_{\text{in}}}$ denotes the weight of a linear layer, 
$\mathbf{U} \in \mathbb{R}^{d_{\text{out}} \times d_{\text{out}}}$ and 
$\mathbf{V} \in \mathbb{R}^{d_{\text{in}} \times d_{\text{in}}}$ are orthogonal matrices containing the left and right singular vectors 
$\mathbf{u}_i \in \mathbb{R}^{d_{\text{out}}}$ and $\mathbf{v}_i \in \mathbb{R}^{d_{\text{in}}}$, respectively, 
$\mathbf{\Sigma} \in \mathbb{R}^{d_{\text{out}} \times d_{\text{in}}}$ is a diagonal matrix whose diagonal entries $\sigma_i$ are singular values in descending order, 
and $R$ is the rank of $\mathbf{WC}$, \emph{i.e.}, $R \leq \min \{ d_{\text{out}}, d_{\text{in}} \}$.

To ensure that the initialization of fine-tuning does not alter the inference output of the pre-trained model, we reconstruct $\mathbf{W}$ as:
\begin{equation}
    \hat{\mathbf{W}} = \text{SVD}(\mathbf{WC})\mathbf{C}^{-1} = \mathbf{U} \mathbf{\Sigma} (\mathbf{V}^{T} \mathbf{C}^{-1}) 
    = \sum_{i=1}^{R} \sigma_i \mathbf{u}_i \hat{\mathbf{v}}_i^{T},
\label{Eq4-1-2}
\end{equation}
where $\mathbf{C}^{-1}$ is the inverse of $\mathbf{C}$, and $\hat{\mathbf{v}}_i^{T}$ denotes the $i$-th row of $\mathbf{V}^{T}\mathbf{C}^{-1}$. 
If $\mathbf{C}$ is not invertible, we enforce invertibility by adaptively adding positive values to its diagonal. 
Specifically, we compute the average of the diagonal entries of $\mathbf{C}$, multiply it by a positive scaling factor, and add this term to the diagonal of $\mathbf{C}$.
This process is repeated until the $\ell_{2}$ distance between $\mathbf{CC}^{-1}$ and the identity matrix falls below a small number, satisfying invertibility. 

Such decomposition leads to a safety-sensitive subspace, because the \emph{leading} components $(\mathbf{u}_i,\hat{\mathbf{v}}_i)$ with large $\sigma_i$ correspond to the safety-relevant directions that are crucial for the safety behaviors against harmful inputs, whereas the bottom components contribute negligibly. 
Accordingly, during fine-tuning, we freeze the safety-relevant components to preserve the safety behaviors they provide. 
Meanwhile, we split out the safety-irrelevant components with the smallest $r$ singular values, which naturally compose two low-rank adapters as:
\begin{equation}
    \mathbf{B} = \mathbf{U}[:, -r:] \sqrt{\mathbf{\Sigma}[-r:]}, \quad
    \mathbf{A} = \sqrt{\mathbf{\Sigma}[-r:]} (\mathbf{V}^{\top} \mathbf{C}^{-1})[-r:,:], 
\label{Eq4-1-3}
\end{equation}
where $\mathbf{B} \in \mathbb{R}^{d_{\text{out}} \times r}$ and $\mathbf{A} \in \mathbb{R}^{r \times d_{\text{in}}}$ are low-rank adapters, $[:, -r:]$ refers to the last $r$ column vectors, $[-r:,:]$ denotes the last $r$ row vectors, and $\sqrt{\mathbf{\Sigma}[-r:]}$ represents a diagonal matrix containing the square root of the smallest $r$ singular values in $\mathbf\Sigma$. 

When fine-tuning on a new dataset, we use $\mathbf{A}$ and $\mathbf{B}$ as the initialized learnable adapters. 
Their product $\mathbf{BA} = \sum_{i=R-r+1}^{R} \sigma_i \mathbf{u}_i \hat{\mathbf{v}}_i^{\top}$ correspond to the pre-trained model weights whose safety-relevant part has been peeled off. 
Compared with LoRA \citep{hulora} that uses zero-initialized adapters, starting from $\mathbf{BA}$ to learn a new task helps to achieve better fine-tuning performance, while maintaining the original safety alignment. 




\subsection{Optimization in harmful-resistant null space}

We begin by recalling the background of null space. 
For two matrices $\mathbf{D}$ and $\mathbf{E}$, the condition $\mathbf{ED}=\mathbf{0}$ implies that each row of $\mathbf{E}$ lies in the left null space of $\mathbf{D}$ \citep{wang2021training,fang2025alphaedit}. After fine-tuning, the update of learnable adapters will inevitably alter the output activations, which may deviate from the original safe output on harmful prompts and undermine the safety mechanism. 
To this end, we further introduce a harmful-resistant null space and project learnable adapters onto this null space such that the weight update will be nullified on harmful inputs. 
Specifically, based on the same safety-triggering prompts, we perform SVD on the covariance matrix of each linear layer as:
\begin{equation}
    \text{SVD}(\mathbf{C}) = \mathbf{Q} \mathbf{\Lambda} \mathbf{Q}^{\top}, 
\label{Eq4-2-2}
\end{equation}
where $\mathbf{Q}\in\mathbb{R}^{d_{\text{in}}\times d_{\text{in}}}$ contains the eigenvectors of $\mathbf{C}$, $\mathbf{\Lambda} = \text{diag}(\lambda_i), 1\le i \le d_{\text{in}},$ represents the eigenvalues of $\mathbf{C}$, and we have $\lambda_i \ge0$ since $\mathbf{C}$ is symmetric positive semi-definite.

We then discard the eigenvectors associated with non-zero eigenvalues. 
On this basis, we construct a projection matrix $\mathbf{P}$ formulated as: 
\begin{equation}
    \mathbf{P} = \hat{\mathbf{Q}}\hat{\mathbf{Q}}^{\top}, 
\label{Eq4-2-3}
\end{equation}
where $\hat{\mathbf{Q}}$ denotes the eigenvectors whose corresponding eigenvalues are zero.  

Since the projector $\mathbf{P}$ can map a matrix into the null space of $\mathbf{C}$, and $\mathbf{C}=\mathbf{X}\mathbf{X}^T$ shares the same null space as the harmful input $\mathbf{X}$, we integrate $\mathbf{P}$ on the product of adapters, \emph{i.e.,} $\mathbf{BA}$.
As a result, $\mathbf{BA}$ is mapped into the null space of $\mathbf{X}$. 
Accordingly, we adjust the frozen part of weight components as 
\begin{equation}
    \mathbf{W}' = \mathbf{W} - \mathbf{BAP}, 
\label{Eq4-2-4}
\end{equation}
to ensure that the initialization of fine-tuning does not alter the inference output of the pre-trained model. 
Consequently, we obtain that:
\begin{equation}
 (\mathbf{W}' + \mathbf{B^*A^*} \cdot \mathbf{P})\mathbf{X} =\mathbf{W}'\mathbf{X},\quad \mathbf{X}\in \mathcal H,
 \label{eq_invariant}
\end{equation}
where $\mathbf{B^*}$ and $\mathbf{A^*}$ refer to the optimized adapter weights after fine-tuning, and $\mathcal H$ denotes the set of harmful prompts. 
Eq. (\ref{eq_invariant}) implies that for harmful inputs $\mathbf{X}$ from the safety-triggering prompts, the output activation remains invariant under adapter updates, thereby preserving the safety behaviors of the original model. In practice, if the harmful prompt space $\mathcal H$ covers sufficient patterns of malicious purposes, the harmful-resistant null space is expected to generalize to unseen harmful data. In experiments, we conduct analysis about the influence of sampling datasets and data size. 

Together, the safety-sensitive subspace and the harmful-resistant null space compose our framework, GuardSpace, as a firm barrier to preserve safety alignment in pre-trained models. 
The pseudo-code of our method is provided in Algorithm \ref{alg 1}.


\begin{algorithm}[t]
   \caption{Overall algorithm of GuardSpace.}
   \label{alg 1}
\begin{algorithmic}[1]
   \STATE {\bfseries Input:} The prompt of harmful dataset, \emph{e.g.,} AdvBench, the downstream task dataset $\mathcal{D}$, the number of layers to optimize $L$, the number of training epochs $T$;
   \STATE {\bfseries Output:} $\mathbf{W}'$, $\mathbf{A}^*$,$\mathbf{B}^*$;
    
    \FOR{$l=1$ {\bfseries to} $L$} 
         \STATE For each layer, initialize $\mathbf{A}$,$\mathbf{B}$ through Eq. (\ref{Eq4-1-1}),Eq. (\ref{Eq4-1-2}) and Eq. (\ref{Eq4-1-3}); 
          \STATE The null space mapping matrix $\mathbf{P}$ of each layer is obtained by Eq.~(\ref{Eq4-2-2}) and Eq.~(\ref{Eq4-2-3});
          \STATE For each layer, obtain $\mathbf{W}'$   through Eq.~(\ref{Eq4-2-4}); 
   \ENDFOR \\
   \FOR{$t=1$ {\bfseries to} $T$}
   \STATE Perform forward propagation through $(\mathbf{W}' + \mathbf{BA}\mathbf{P})\mathbf{X} $ for each layer and optimize $\mathbf{A}$ and $\mathbf{B}$ using the supervised fine-tuning loss on $\mathcal{D}$.
 
   \ENDFOR 
   
\end{algorithmic}

\end{algorithm}


\section{Experiments}

\subsection{Setup}
\label{sec:setup}
\textbf{Datasets.} Prior studies indicate that adapting a model via fine-tuning can introduce significant safety risks: even limited exposure to adversarial or seemingly benign samples during training may erode built-in safeguards, yielding unsafe generations after adaptation \citep{huangbooster, bianchisafety, qi2024fine}. To emulate harmful fine-tuning attacks, and following the setup in \cite{yang2025asft}, we employ three tasks, SST-2 \citep{socher2013recursive}, AGNEWS \citep{zhang2015character} and GSM8K \citep{cobbe2110training}, as our fine-tuning targets. Detailed details for constructing the training datasets are provided in Appendix \ref{sec:data_setting}.

\textbf{Base LLMs.} We assess our approach on three instruction-tuned LLMs: Llama-2-7B-Chat \citep{touvron2023llama}, Gemma-2-9B-IT \citep{team2024gemma} and Qwen-2-7B-Instruct \citep{hui2024qwen2}. The download links for the models and datasets are provided in Appendix \ref{App_Data_Model}.

\textbf{Baseline Methods.} Beyond LoRA, we benchmark seven defensive baselines–SafeInstr \citep{bianchisafety}, BEA \citep{wang2024backdooralign}, Lisa in two variants (base and aligned) \citep{huang2024lazy}, Safe-LoRA \citep{hsusafe}, SaLoRA \citep{lisalora}, and AsFT \citep{yang2025asft}.  Comprehensive method summaries and configuration settings are provided in Appendix \ref{App_Baseline}. 

\textbf{Evaluation Metrics and Settings.}
In line with prior work \citep{huangbooster, huangantidote, yang2025asft}, we adopt two evaluation metrics, both computed on the fine-tuned model. \textit{(1)} Fine-tuning Accuracy (FA): 
Top-1 accuracy of the model on the held-out test set for the corresponding fine-tuning task. \textit{(2)} Harmfulness Score (HS): Following \cite{ji2023beavertails}, we apply a moderation classifier to the model’s responses to previously unseen malicious prompts; HS is the proportion of outputs flagged as unsafe.
In this section, we demonstrate that \method preserves safety alignment while simultaneously improving downstream task utility. Unless otherwise stated, we estimate the safety-sensitive subspace using 520 prompts from AdvBench \citep{zou2023universal}, and construct the harmful-resistant null space projector \(\mathbf{P}\) by randomly sampling 520 prompts from RealToxicityPrompts \citep{xie2024sorrybench}. For adapter initialization, we allocate trainable capacity to the safety-irrelevant components with the smallest $128$ singular values, in accordance with Eq.~(\ref{Eq4-1-3}).

\subsection{Main results}


\textbf{Generalization on fine-tuning datasets.}
As shown in Tab.\ref{tab:ftdatasets}, we fine-tune Llama-2-7B-Chat on SST-2, AGNEWS, and GSM8K, reporting safety (HS$\downarrow$) and utility (FA$\uparrow$). Although LoRA improves FA, it degrades the safety performance a lot on all three datasets.  Existing safe-related methods can achieve the better safety performance than LoRA usually with the acceptable task performance, where AsFT is strongest among them but remains above base safety (avg HS $8.13\%$ vs.\ $2.40\%$). \method matches base-level safety (avg HS $2.40\%$) while raising utility ($+33.95\%$). These results indicate that our method addresses this limitation: existing fine-tuning defenses seldom identify safety-relevant components or harmful update directions. By explicitly isolating the former and constraining the latter, \method reduces training conflicts and achieves a more favorable balance between safety preservation and task performance. On SST-2, our method even achieves better HS than base model, which we attribute to the fixed null-space projector limiting first-order effects of adapter updates on harmful inputs.

\begin{table*}[t]
\centering
\scriptsize 

\setlength{\tabcolsep}{3pt}      
\renewcommand{\arraystretch}{1.00} 
\setlength{\aboverulesep}{0pt}    
\setlength{\belowrulesep}{0pt}    
\setlength{\abovetopsep}{0pt}     
\setlength{\belowbottomsep}{0pt}  
\setlength{\cmidrulekern}{0.3em}  

\setlength{\tabcolsep}{5pt}
\renewcommand{\arraystretch}{1.05}
\caption{\small{Performance of \textbf{Llama-2-7B-Chat} fine-tuned on different datasets. HS$\downarrow$ indicates lower is better; FA$\uparrow$ indicates higher is better. Best results are shown in bold; second-best results are underlined.}}
\label{tab:ftdatasets}
\resizebox{\linewidth}{!}{
\begin{tabular}{l *{4}{cc}}
\toprule
\multirow{2}{*}{\textbf{Methods} (\textit{Llama-2-7B-Chat})} &
\multicolumn{2}{c}{\textbf{SST2}} &
\multicolumn{2}{c}{\textbf{AGNEWS}} &
\multicolumn{2}{c}{\textbf{GSM8K}} &
\multicolumn{2}{c}{\textbf{Average}} \\
\cmidrule(lr){2-3}\cmidrule(lr){4-5}\cmidrule(lr){6-7}\cmidrule(lr){8-9}
& \textbf{HS} $\downarrow$ & \textbf{FA} $\uparrow$
& \textbf{HS} $\downarrow$ & \textbf{FA} $\uparrow$
& \textbf{HS} $\downarrow$ & \textbf{FA} $\uparrow$
& \textbf{HS} $\downarrow$ & \textbf{FA} $\uparrow$ \\
\midrule
Base Model  & 2.40 & 28.90 & 2.40 & 64.70 & 2.40 & 13.80 & 2.40 & 35.80 \\
\midrule
LoRA \citep{hulora}                & 48.00 & 94.50 & 17.60 & 84.30 & 56.00 & 23.80 & 40.53 & 67.53 \\
Lisa-base \citep{huang2024lazy}    & 27.60 & \textbf{96.90} & 27.20 & 73.50 & 35.20 & 24.00 & 30.00 & 64.80 \\
Lisa-aligned \citep{huang2024lazy} &  \underline{5.60} & 93.58 & 16.80 & 81.80 & 16.00 & 19.40 & 12.80 & 64.93 \\
SafeInstr \citep{bianchisafety}    &  \,9.20 & 93.35 & 16.80 & 84.30 & 17.60 & 19.30 & 14.53 & 65.65 \\
BEA \citep{wang2024backdooralign}  &  \,7.20 & 91.63 & 16.40 & \underline{84.40} & 38.80 & 21.00 & 20.80 & 65.68 \\
Safe LoRA \citep{hsusafe}      & 11.20 & 89.24 &  \,5.60 & 81.20 & 36.00 & 23.60 & 17.60 & 64.68 \\
AsFT \citep{yang2025asft}          &  \,6.00 & 93.32 & \underline{4.00} & 84.30 & \underline{14.40} & \underline{26.00} &  \underline{8.13} & \underline{67.87} \\
\rowcolor{gray!12} \method (Ours)        &  \,\textbf{1.20} & \underline{95.64} & \textbf{2.40} & \textbf{85.60} & \textbf{3.60} & \textbf{28.00} & \textbf{2.40} & \textbf{69.75} \\
\bottomrule
\end{tabular}
}
 \vspace{-1.5em}
\end{table*}

\textbf{Generalization to models.}
We assess cross-model generalization by fine-tuning three architectures (Llama-2-7B-Chat, Qwen-2-7B-Instruct, Gemma-2-9B-IT) on GSM8K and then evaluating safety and utility. As shown in Tab.\ref{tab:model_gsm8k}, LoRA yields unsurprising high FA and large HS (e.g., HS $30.00\%$ on Qwen-2-7B).  Prior defense reduce HS but show mixed FA across models. \method attains the lowest HS on all three models while keeping FA competitive, improving both HS and FA on Llama-2-7B and achieving the lowest HS with near-top FA on Qwen-2-7B. Averaged over models, \method reaches HS $3.20\%$ and FA $54.53\%$, consistent with our design of safety-sensitive initialization and null-space constrained updates. Note on Gemma-2-9B-IT, the base model exhibits higher FA than several fine-tuned variants. We attribute this to its strong instruction tuning on reasoning-style data (good zero-shot CoT calibration), coupled with limited-task fine-tuning that can perturb internal reasoning features or overfit to small supervision. Despite this, \method attains the lowest HS while maintaining competitive FA on Gemma-2-9B-IT.


\begin{table*}[t]
\scriptsize 

\setlength{\tabcolsep}{3pt}      
\renewcommand{\arraystretch}{1.00} 
\setlength{\aboverulesep}{0pt}    
\setlength{\belowrulesep}{0pt}    
\setlength{\abovetopsep}{0pt}     
\setlength{\belowbottomsep}{0pt}  
\setlength{\cmidrulekern}{0.3em}  

\renewcommand{\arraystretch}{1.0}
\caption{\small{Performance of different model architectures on {GSM8K}. HS$\downarrow$ (lower is better); FA$\uparrow$ (higher is better).}}
\vspace{-2mm}
\label{tab:model_gsm8k}
\resizebox{\textwidth}{!}{%
\begin{tabular}{l cc cc cc cc}
\toprule
\multirow{2}{*}{\textbf{Methods}} &
\multicolumn{2}{c}{\textbf{Llama-2-7B-Chat}} &
\multicolumn{2}{c}{\textbf{Qwen-2-7B-Instruct}} &
\multicolumn{2}{c}{\textbf{Gemma-2-9B-IT }} &
\multicolumn{2}{c}{\textbf{Average}} \\
\cmidrule(lr){2-3}\cmidrule(lr){4-5}\cmidrule(lr){6-7}\cmidrule(lr){8-9}
& \textbf{HS$\downarrow$} & \textbf{FA$\uparrow$}
& \textbf{HS$\downarrow$} & \textbf{FA$\uparrow$}
& \textbf{HS$\downarrow$} & \textbf{FA$\uparrow$}
& \textbf{HS$\downarrow$} & \textbf{FA$\uparrow$} \\
\midrule

Base Model                        & 2.40 & 13.80 &  4.80 &  49.00 &  2.00 & 77.20 & 3.07 & 46.70 \\
\midrule
LoRA  \citep{hulora}                      & 56.00 & 23.80 & 30.00 & \textbf{66.40} & 50.00 & 69.80 & 45.33 & 53.33 \\
SafeInstr \citep{bianchisafety} & 17.60 & 19.30 &  7.20 & 63.00 &  \underline{2.80} & \underline{76.20} &  9.20 & 52.83 \\
BEA  \citep{wang2024backdooralign} & 38.80 & 21.00 &  8.40 & 54.60 &  4.80 & 65.00 & 17.33 & 46.87 \\
Safe LoRA \citep{hsusafe}         & 36.00 & 23.60 & 10.40 & 50.40 &  6.00 & \textbf{77.00} & 17.47 & 50.33 \\
AsFT \citep{yang2025asft}         & \underline{14.40} & \underline{26.00} & \underline{7.20} & 63.40 & 4.80 & 74.20 & \underline{8.80} & \underline{54.53} \\
\rowcolor{gray!12} \method (Ours) & \textbf{3.60} & \textbf{28.00} & \textbf{3.20} & \underline{65.40} & \textbf{2.80} & 70.20 & \textbf{3.20} & \textbf{54.53} \\
\bottomrule
\end{tabular}%
}
\vspace{1mm}
\end{table*}

\textbf{Robustness against varying ratios of unsafe examples.}
We fine-tune Llama-2-7B-Chat on GSM8K with an unsafe proportion
$p\in\{0,0.05,0.10,0.15,0.20\}$ and report HS/FA in
Tab.\ref{tab:gsm8k_harm_ratio}. As $p$ increases, most baselines show clear safety drift. Among them, LoRA has the most  significant decline trend, whose HS rises from $8.80\%$
at clean dataset to $60.00\%$ at $p{=}0.20$ (FA stays near $24{\sim}27\%$). Although safety-oriented methods can alleviate this trend, they still produce worse HS with a larger $p$. For example, AsFT has HS
$2.40\%$ when clean yet reaches $20.80\%$ at $p{=}0.20$. In contrast, \method keeps HS uniformly low across all $p$, and achieves the highest average FA ($25.88\%$). Overall,
\method maintains near-floor harmfulness while retaining utility under
up to $20\%$ poisoning, proving the effectiveness of sensitive initialization and harmful-resistant null space constraint.

\begin{table}[t]
\scriptsize 

\setlength{\tabcolsep}{3pt}      
\renewcommand{\arraystretch}{1.00} 
\setlength{\aboverulesep}{0pt}    
\setlength{\belowrulesep}{0pt}    
\setlength{\abovetopsep}{0pt}     
\setlength{\belowbottomsep}{0pt}  
\setlength{\cmidrulekern}{0.3em}  

\caption{\small{Performance of Llama-2-7B-Chat on GSM8K under varying unsafe ratios.}}
\vspace{-2mm}
\label{tab:gsm8k_harm_ratio}
\footnotesize                           
\setlength{\tabcolsep}{4pt}            
\renewcommand{\arraystretch}{1.08}     
\resizebox{\linewidth}{!}{             
\begin{tabular}{l|cccccc|cccccc}
\toprule
\multirow{2}{*}{Methods} &
\multicolumn{6}{c|}{\textbf{Harmful Score} $\downarrow$} &
\multicolumn{6}{c}{\textbf{Finetune Accuracy} $\uparrow$} \\
\cmidrule(lr){2-7}\cmidrule(lr){8-13}
& clean & $p=0.05$ & $p=0.10$ & $p=0.15$ & $p=0.20$ & Avg. 
& clean & $p=0.05$ & $p=0.10$ & $p=0.15$ & $p=0.20$ & Avg. \\
\midrule
LoRA         &  8.80 & 40.80 & 56.00 & 34.00 & 60.00 & 39.92  & 24.60 & \underline{27.20} & 23.80 & 22.40 & \underline{24.60} & \underline{24.52} \\
Lisa-base    & 39.60 & 32.80 & 35.20 & 29.60 & 31.20 & 33.68  & 20.40 & 19.80 & 24.00 & 21.60 & 20.80 & 21.32 \\
Lisa-aligned & 14.40 & 16.00 & 16.00 & 21.60 & 23.60 & 18.32  & 20.00 & 20.60 & 19.40 & 19.80 & 24.40 & 20.84 \\
SafeInstr    &  5.20 & 13.20 & 17.60 & 37.20 & 43.60 & 23.36  & 20.50 & 22.40 & 19.30 & 22.10 & 20.50 & 20.96 \\
BEA          &  6.40 & 32.80 & 38.80 & 32.80 & 38.00 & 29.76  & 21.60 & 21.60 & 21.00 & 20.00 & 20.00 & 20.84 \\
Safe LoRA    &  8.80 & 22.80 & 36.00 & 33.20 & 40.80 & 28.32  & \underline{24.60} & 22.60 & 23.60 & \textbf{24.20} & 24.00 & 23.80 \\
AsFT   &  \textbf{2.40} & \underline{7.20} & \underline{14.40} & \underline{15.80} & \underline{20.80} & \underline{12.12}& 23.20 & 24.20 & \underline{26.00} & 23.20 & \textbf{24.80} & 24.28 \\

\rowcolor{gray!15} Ours &  \underline{2.80} & \textbf{1.20} & \textbf{3.60}  &\textbf{2.80} & \textbf{2.40} & \textbf{2.56}
             & \textbf{26.00} & \textbf{28.60} & \textbf{28.00} & \underline{22.40} & 24.40 &\textbf{25.88}  \\
\bottomrule
\end{tabular}
} 
\end{table}

\subsection{Ablation Studies and Analysis}

\textbf{Effectiveness of safety-sensitive subspace and harmful-resistant null space.}
Tab.\ref{tab:ablation-acc-hs} examines the contribution of each component on Llama-2-7B-Chat (GSM8K). Removing the subspace initialization (``w/o subspace initialization'') raises HS from $3.60\%$ to $5.20\%$ ($+1.60\%$) with only a marginal FA change ($28.00\%\!\to\!26.20\%$), indicating that initializing from the safety-irrelevant components with the smallest 
$r$ singular values  improves safety at little utility cost.” In contrast, removing the null space projector (``w/o null space projector'') preserves or slightly boosts FA ($28.60\%$) but causes a drastic safety collapse (HS $3.60\%\!\to\!52.00\%$, $\sim$14.40$\times$), showing that the projector is the primary driver of safety preservation. Together, the two parts yield the best safety–utility balance: the subspace initialization step places trainable capacity in safety-insensitive directions and trims harmfulness without sacrificing accuracy, while the projector prevents harmful activation shifts. 

Fig.\ref{fig:low_r_ASR} in Appendix \ref{sec:App_safte_sen_null} evaluates the \emph{safety-sensitive initialization}.
We reconstruct $\hat{\mathbf{W}}$ by Eq.~(\ref{Eq4-1-2}) and discard the trailing
$r\!\in\!\{0,16,32,64,128,256,512,1024\}$ safety-irrelevant components.
Using AdvBench prompts \citep{yuan2023asvd}, we compare Plain SVD, ASVD, and Ours.
Plain SVD collapses at large $r$ (ASR spikes) and ASVD drifts,
whereas ours keeps ASR low ($1.82$–$5.15\%$) across all $r$,
indicating that covariance preconditioning concentrates safety-relevant structure
in the retained components so that $\hat{\mathbf{W}}$ preserves refusals without further training.

Fig.\ref{fig:null_space_projector} in Appendix \ref{sec:App_safte_sen_null} tests the \emph{harmful-resistant null space}.
We train adapters with or without the fixed projector $\mathbf{P}$ and report HEx-PHI ASR over epochs.
With $\mathbf{P}$, ASR remains near floor from epochs $1$–$8$, consistent with
$(\mathbf{W}' + \mathbf{B^*A^*} \cdot \mathbf{P})\mathbf{X} =\mathbf{W}'\mathbf{X}$ keeping harmful activations unchanged during learning.
Without $\mathbf{P}$, ASR surges after 6–7 epochs ($>\!20\%$), revealing drift along unsafe directions.
Thus, the projector constrains updates within the harmful-resistant null space and complements the initialization: the model is safe at step~0, and safety is maintained throughout training.

\begin{table}[t]
\renewcommand{\arraystretch}{0.8}
\centering
\vspace{-2mm}
\caption{\small{The ablation study of \method with {Llama-2-7B-Chat} on GSM8K.}}
\vspace{-2mm}
\label{tab:ablation-acc-hs}
\begin{tabular}{l c c c c}
\toprule
\textbf{Methods} & \textbf{Adapter Initialization} & \textbf{Projector} & \textbf{HS}\,$\downarrow$ & \textbf{FA}\,$\uparrow$ \\
\midrule
LoRA                    & zero                    & no   & 56.00 & 23.80 \\
w/o subspace initialization & zero                    & null space & 5.20  & 26.20 \\
w/o null space projector& safety-irrelevant subspace & no   & 52.00 & \textbf{28.60} \\
\rowcolor{gray!12}
\method(Ours)         & safety-irrelevant subspace & null space & \textbf{3.60} & 28.00 \\
\bottomrule
\end{tabular}
\end{table}

\begin{figure}[t]
  \centering

  \begin{subfigure}[b]{0.495\textwidth}
    \includegraphics[width=\linewidth]{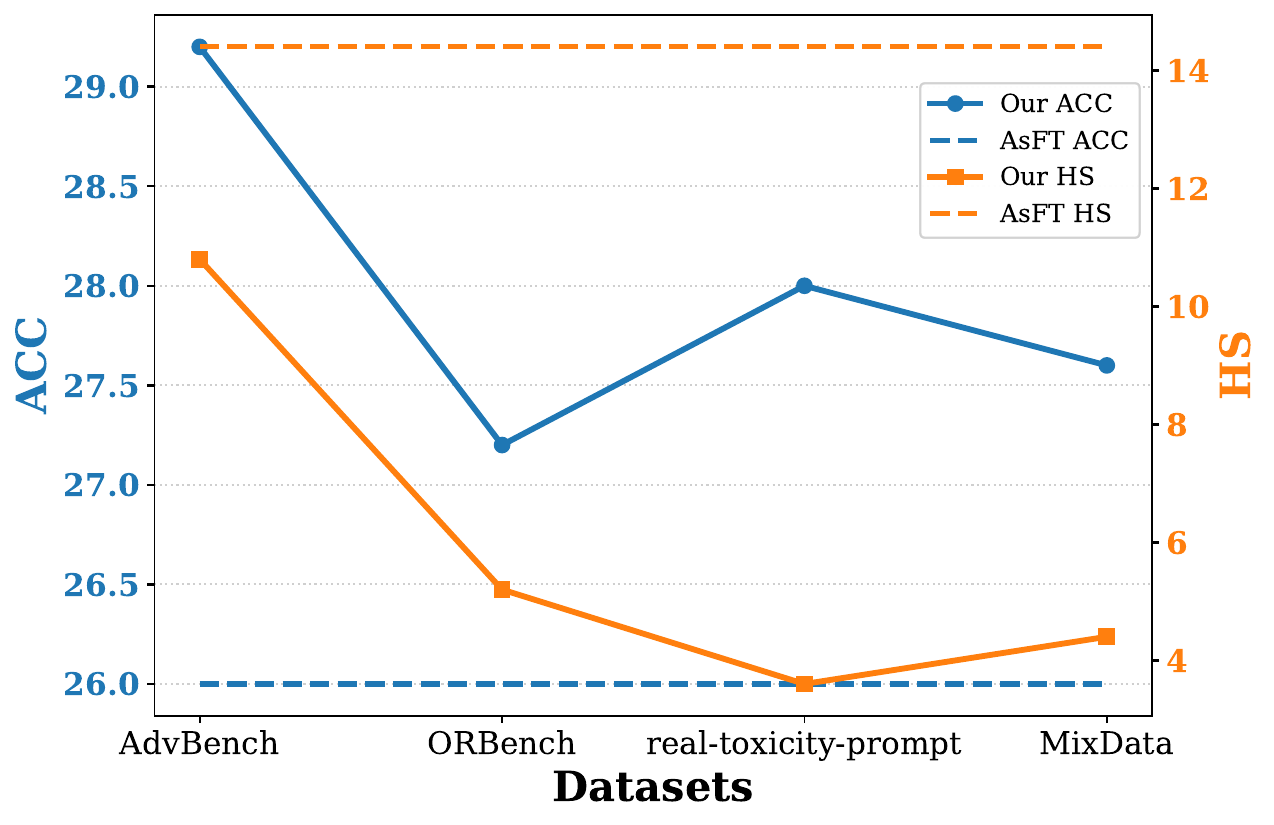}
    \caption{ \small{Different harmful-datasets }} 
    \label{fig:dataset_choisse}
  \end{subfigure}
   \hfill
  \begin{subfigure}[b]{0.495\textwidth}
    \includegraphics[width=\linewidth]{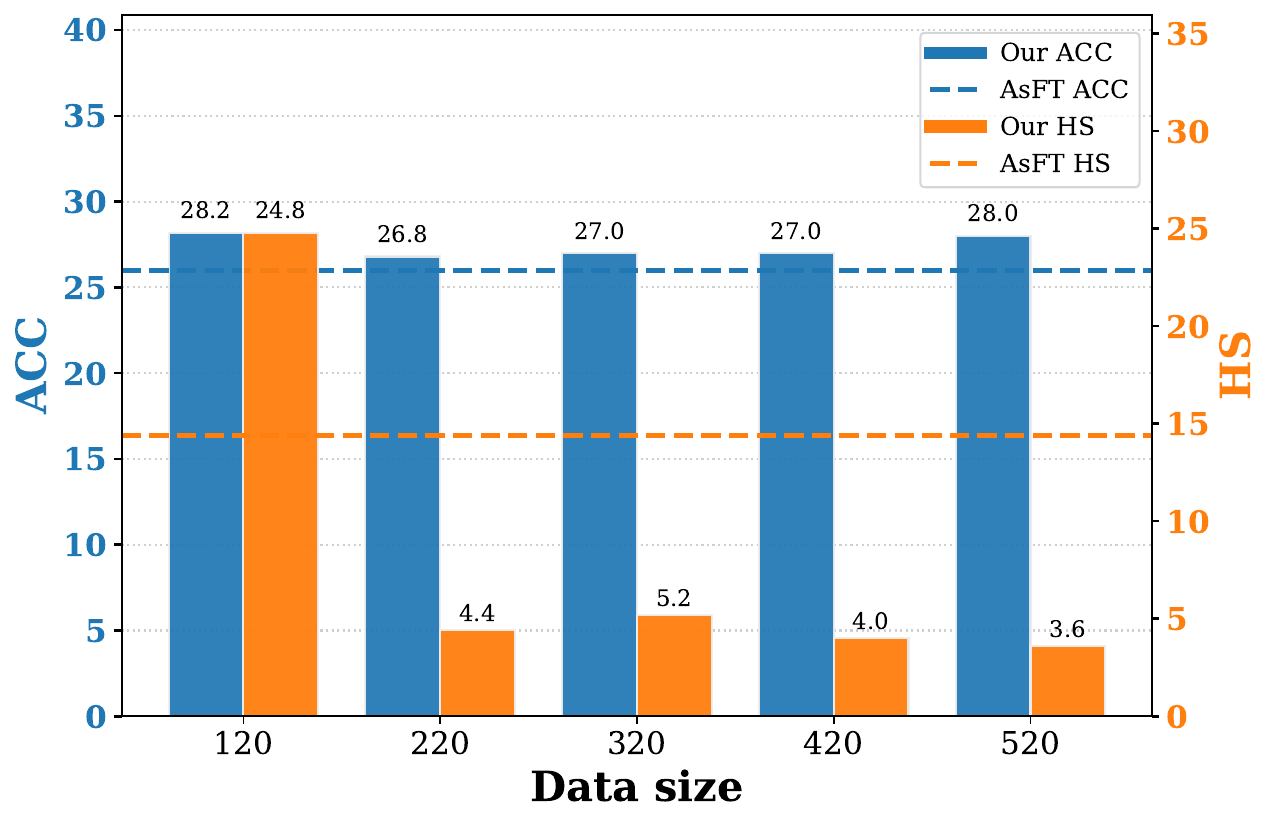}
    \caption{\small{ The size of harmful data }} 
    \label{fig:datasize}
  \end{subfigure}
  \hfill
  
   \vspace{-6mm}
   \caption{\small{GuardSpace Null-Space Projector Analysis on Llama-2-7B-Chat (GSM8K)
(a) Effect of harmful-dataset choice on GuardSpace’s null space projector for Llama-2-7B-Chat (GSM8K); (b) Effect of data size on GuardSpace’s null-space projector for Llama-2-7B-Chat (GSM8K)}}

 \vspace{-4mm}
  \label{fig:datasets_and_dataseize}
\end{figure}

\textbf{Influence of sampling dataset and data size.}
To test robustness to the harmful corpus used for null  space projector construction, we estimate the covariance $\mathbf{C}$ from hidden activations elicited by 520 safety-triggering prompts drawn from AdvBench, ORBench, RealToxicityPrompt, or their equal-mix (MixData), build the fixed projector $\mathbf{P}$, and fine-tune Llama-2-7B-Chat on GSM8K.
Fig.\ref{fig:dataset_choisse} shows downstream utility (ACC; left axis) and harmfulness (HS$\downarrow$; right axis), with AsFT as a reference.
Across all corpora, GuardSpace achieves higher ACC and lower HS than AsFT, indicating that the \emph{harmful-resistant null space} learned from a few hundred prompts generalizes well.
Dataset identity causes only mild variation: AdvBench gives the highest ACC, RealToxicityPrompt has the lowest HS, and MixData provides a balanced trade-off. Fig.\ref{fig:datasize} varies the number of harmful prompts used to build $\mathbf{P}$. We can find that, once the size reaches $\ge\!220$, HS in ours falls from ${24.8\%}$ at $120$ to ${4}$–${6\%}$  and then plateaus; downstream accuracy in ours remains stable  across all sizes. Compared with AsFT (dashed lines), GuardSpace consistently yields much lower HS with comparable or slightly higher ACC.
Thus, $200$–$300$ prompts suffice to learn a robust projector that preserves safety without sacrificing utility, with larger sets offering diminishing returns.

\textbf{Influence of adapter rank.}
We also analyze the influence of adapter rank and provide the results in Appendix~\ref{sec:App_adapter_rank}.

\section{Conclusion}

We introduced GuardSpace, a fine-tuning framework that preserves safety alignment while retaining downstream utility via two parts: a safety-aware initialization (covariance-preconditioned factorization that allocates trainable capacity to safety-irrelevant directions) and a harmful-resistant null-space projector (constrains adapter updates so harmful activations remain unchanged). GuardSpace lowers harmfulness while maintaining or improving accuracy across sentiment, topic classification, and math reasoning. It matches base-level safety on Llama-2-7B-Chat while outperforming LoRA and prior defenses on utility, generalizes across Llama-2-7B-Chat, Llama-3-8B-Instruct and Gemma-2-9B-IT with near-floor HS and competitive FA, sustains low HS with up to $20\%$ unsafe data; the projector is the main safety driver, with initialization providing smaller gains at minimal utility cost.

\bibliography{iclr2026_conference}
\bibliographystyle{iclr2026_conference}

\newpage
\appendix
\section{Experiment Details}

\subsection{Poisoning Setup and Dataset Composition}
\label{sec:data_setting}
To ensure a fair comparison, we follow the experimental setup of AsFT \citep{yang2025asft}.  For each configuration, we construct a training set of 1000 examples by mixing a proportion $p$ of unsafe (poisoned) examples from the Harmful corpus \citep{yuan2025towards} with the complementary proportion $1-p$ of benign examples (i.e., $1000p$ unsafe and $1000(1-p)$ benign). 
 Unless otherwise specified, we fix $p=0.1$.

\subsection{Dataset and Model}
\label{App_Data_Model}
\begin{table}[h!]
\caption{\small{Datasets (top) and models (bottom) with sources.}}
\centering
\small
\begin{tabular}{@{}llp{0.62\linewidth}@{}}
\toprule
\textbf{Type} & \textbf{Name} & \textbf{Source} \\
\midrule
\multirow{7}{*}{Dataset}
& AdvBench & \Src{https://huggingface.co/datasets/walledai/AdvBench} \\
& RealToxicityPrompts & \Src{https://huggingface.co/datasets/sorry-bench/sorry-bench-202406} \\
& OR-Bench & \Src{https://huggingface.co/datasets/bench-llm/or-bench}\\
\midrule
\multirow{6}{*}{Model}
& Llama-2-7B-Chat & \Src{https://huggingface.co/TheBloke/Llama-2-7B-Chat-fp16} \\
& Gemma-2-9B-It & \Src{https://huggingface.co/google/gemma-2-9b-it} \\
&  Qwen-2-7B-Instruct & \Src{https://huggingface.co/Qwen/Qwen2-7B-Instruct} \\
&  beaver-dam-7b & \Src{https://huggingface.co/PKU-Alignment/beaver-dam-7b} \\

\bottomrule
\end{tabular}

\end{table}

\subsection{Baseline summaries and configuration settings.}
\label{App_Baseline}

\paragraph{Baselines.}
We consider seven representative approaches:
\begin{itemize}
  \item \textbf{LoRA} \citep{hulora}. The standard supervised fine-tuning paradigm implemented with Low-Rank Adaptation.
  \item \textbf{Lisa}. A two-stage optimization framework. Lisa-base \citep{huang2024lazy} starts from base models and performs alignment followed by task tuning; Lisa-aligned \citep{huang2024lazy} begins from already aligned models and further tunes on BeaverTails \citep{ji2023beavertails}.
  \item \textbf{SafeInstr} \citep{bianchisafety}. Augments the fine-tuning corpus with carefully curated safety-oriented examples to reinforce safe behavior.
  \item \textbf{BEA} \citep{wang2024backdooralign}. Introduces stealthy trigger prompts as backdoor cues, binding them to safe generations during fine-tuning.
  \item \textbf{Safe LoRA} \citep{hsusafe}. Constrains LoRA parameter updates to subspaces associated with safety-aligned directions and is applied after standard fine-tuning.
  \item \textbf{AsFT} \citep{yang2025asft}. Anchors safety during fine-tuning by using the \emph{alignment direction} (the weight difference between aligned and base models) as a guide; updates orthogonal to this direction are suppressed to keep optimization within a narrow safety basin.
\end{itemize}
Among these, LoRA, Lisa, SafeInstr, BEA and AsFT act during the fine-tuning stage, whereas Safe LoRA is post-hoc. We also attempted to reproduce SaLoRA \citep{lisalora}, but under our experimental setup its results were consistently below all the reported baseline methods. Therefore, SaLoRA is not included in the compared methods.
\paragraph{Implementation details used in our study.}
\begin{itemize}
  \item \textbf{LoRA} \citep{hulora}: We adopt a standard setup with rank $r=8$ and target the attention projection modules q and v. The learning rate is $5\times10^{-5}$, batch size is $8$, and training runs for $10$ epochs. The dataset follows the default mixing strategy, combining harmful data with proportion $p$.
  \item \textbf{Lisa-base} \citep{huang2024lazy}: A two-phase schedule per base model. Phase~1 uses alignment data (e.g., instruction-tuning samples). Phase~2 reuses the same alignment set and adds a proximal term to prevent excessive drift between phases. LoRA hyperparameters match LoRA ($r=8$, q/v, learning rate $5\times10^{-5}$ , batch size $8$, $10$ epochs).
  \item \textbf{Lisa-aligned} \citep{huang2024lazy}: In contrast to Lisa-base, this variant starts from an \emph{aligned/chat} model (e.g., Llama-2-Chat). We then apply only the second phase on BeaverTails \citep{ji2023beavertails} with a proximal constraint on the parameter updates. LoRA hyperparameters mirror LoRA.
  \item \textbf{SafeInstr} \citep{bianchisafety}: We inject safety-enhanced samples amounting to $10\%$ of the harmful portion of the dataset. Other hyperparameters follow the LoRA defaults ($r=8$, q/v, learning rate $5\times10^{-5}$, batch size $8$, and $10$ epochs).
  \item \textbf{BEA} \citep{wang2024backdooralign}: We use the official backdoor samples, also set to $10\%$ of the harmful data. Fine-tuning otherwise matches the LoRA settings ($r=8$, q/v, learning rate $5\times10^{-5}$, batch size $8$, $10$ epochs).
  \item \textbf{Safe LoRA} \citep{hsusafe}: After completing standard LoRA fine-tuning, we insert projection layers that map LoRA updates into safety-aligned subspaces. Following prior settings, we place projections on $40$ layers to achieve a favorable safety–utility trade-off.
  \item \textbf{AsFT} \citep{yang2025asft}: We keep the LoRA schedule unchanged (rank 8 on $q$/$v$, learning rate $5\times10^{-5}$, batch size 8, 10 epochs). AsFT adds a safety regularizer during training that keeps updates aligned with the alignment direction for each layer, defined as the weight difference between an aligned/chat checkpoint and its base counterpart, and penalizes the orthogonal component of each update. The regularization coefficient $\lambda$ is set to 1.
\end{itemize}

\section{Additional Results}
\label{sec:add_analysis}

\subsection{Effectiveness of safety-sensitive subspace and harmful-resistant null space.}
\label{sec:App_safte_sen_null}

\begin{figure}[!htp]
  \centering

  \begin{subfigure}[b]{0.47\textwidth}
    \includegraphics[width=\linewidth]{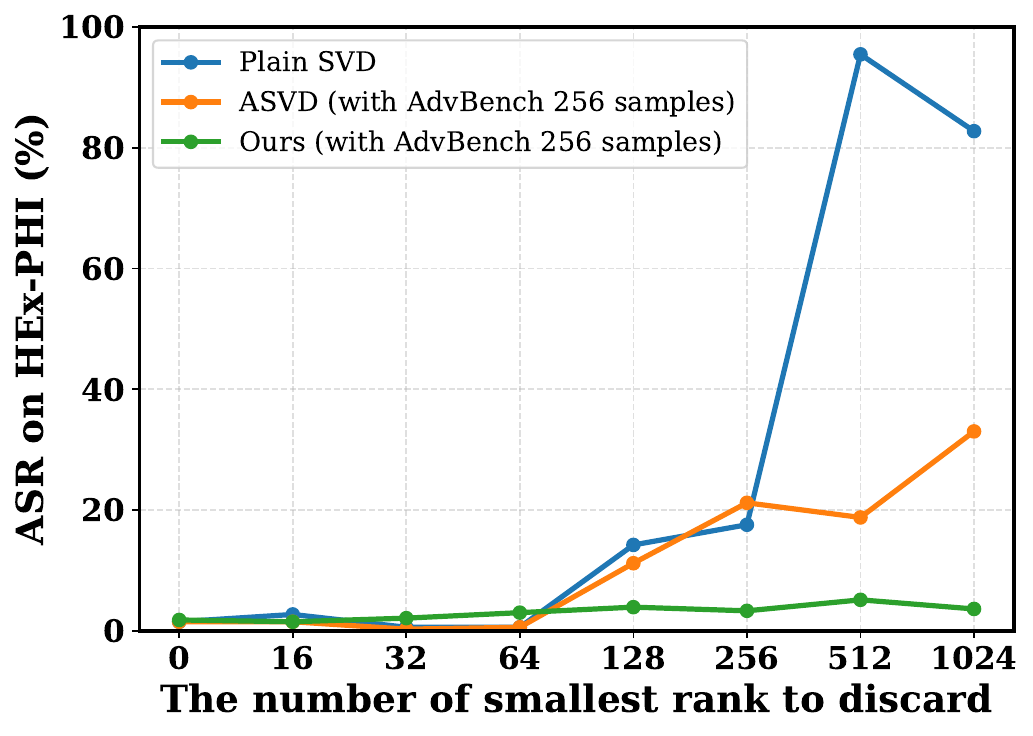}
    \caption{\small{Effect of discarding safety-irrelevant components on ASR.}} 
    \label{fig:low_r_ASR}
  \end{subfigure}
   \hfill
  \begin{subfigure}[b]{0.47\textwidth}
    \includegraphics[width=\linewidth]{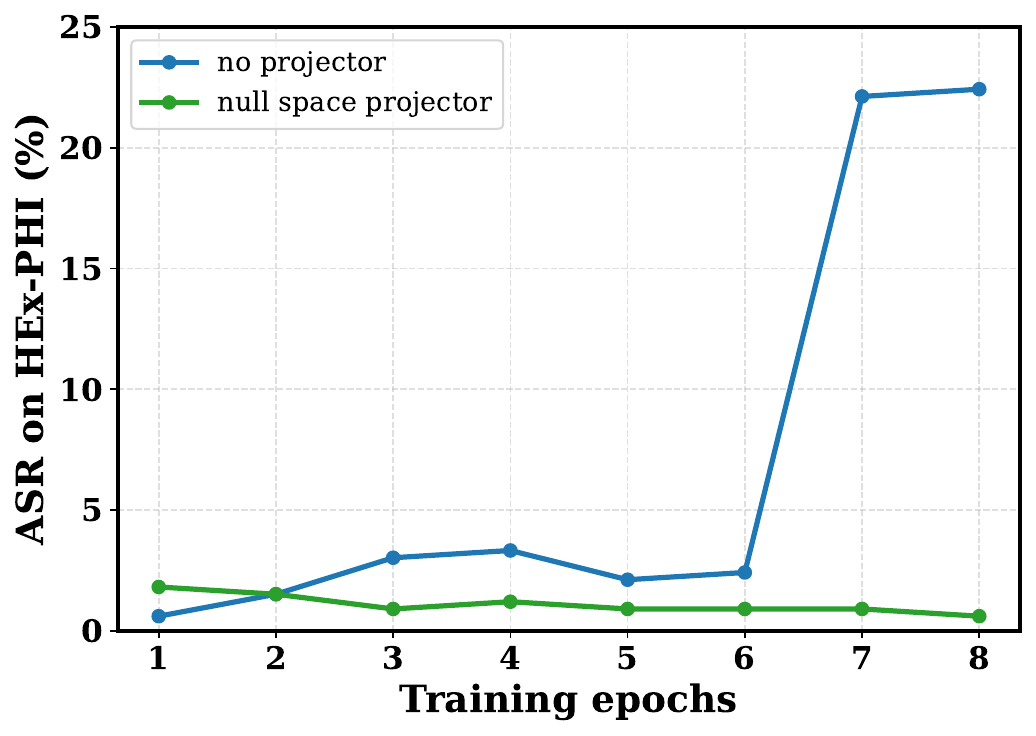}
    \caption{\small{Effect of null space projector vs no prpjector on ASR.}} 
    \label{fig:null_space_projector}
  \end{subfigure}
  \hfill



   \caption{\small{Effect of safety-irrelevant components removal and null space projection on ASR.}}

  \label{fig:tow_stage_validation}
\end{figure}

Fig.\ref{fig:low_r_ASR}  evaluates the \emph{safety-sensitive initialization}.
We reconstruct $\hat{\mathbf{W}}$ by Eq.~(\ref{Eq4-1-2}) and discard the trailing
$r\!\in\!\{0,16,32,64,128,256,512,1024\}$ safety-irrelevant components.
Using AdvBench prompts \citep{yuan2023asvd}, we compare Plain SVD, ASVD, and Ours.
Plain SVD collapses at large $r$ (ASR spikes) and ASVD drifts,
whereas Ours keeps ASR low ($1.82$–$5.15\%$) across all $r$,
indicating that covariance preconditioning concentrates safety-relevant structure
in the retained components so that $\hat{\mathbf{W}}$ preserves refusals without further training.

Fig.\ref{fig:null_space_projector} tests the \emph{harmful-resistant null space}.
We train adapters with or without the fixed projector $\mathbf{P}$ and report HEx-PHI ASR over epochs.
With $\mathbf{P}$, ASR remains near floor from epochs $1$–$8$, consistent with
$(\mathbf{W}' + \mathbf{B^*A^*} \cdot \mathbf{P})\mathbf{X} =\mathbf{W}'\mathbf{X}$ keeping harmful activations unchanged during learning.
Without $\mathbf{P}$, ASR surges after 6–7 epochs ($>\!20\%$), revealing drift along unsafe directions.
Thus, the projector constrains updates within the harmful-resistant null space and complements the initialization: the model is safe at step~0, and safety is maintained throughout training.



\subsection{Influence of adapter rank.}
\label{sec:App_adapter_rank}
\begin{figure}[H]  
  \centering
  \includegraphics[width=0.65\linewidth]{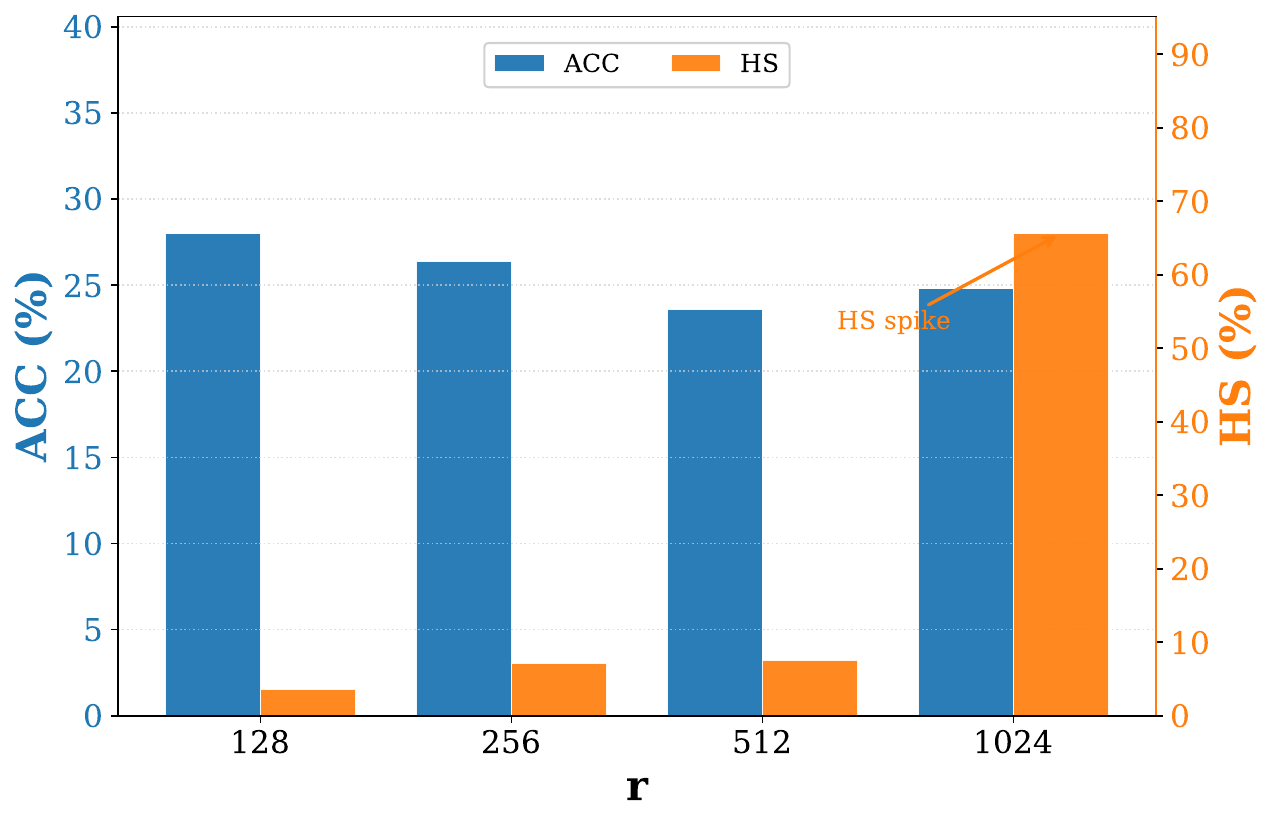}
  \caption{\small{Effect of the number of safety-irrelevant components used for adapter initialization.}}
  \label{fig:rank}
\end{figure}

We explore how many safety-irrelevant components to use for adapter initialization. 
Fig.\ref{fig:rank} reports ACC (left axis) and HS$\downarrow$ (right axis) on GSM8K as we vary $r$. 
Using a small number of safety-irrelevant components ($r{=}128$ -- $512$) keeps HS low while maintaining ACC, 
whereas an overly large $r$ ($1024$) causes an HS spike, suggesting that truncation begins to remove safety-relevant structure.


\end{document}